\RequirePackage{fix-cm}
\documentclass[twocolumn]{svjour3}          
\smartqed  
\usepackage{graphicx}
\usepackage{float}
\usepackage{url}
\usepackage{array,booktabs}
\usepackage{xcolor}

\begin{document}

\title{Seeing Colors: Learning Semantic Text Encoding for Classification}

\titlerunning{Text Classification using Convolutional Neural Networks}        

\author{Shah Nawaz         \and
        Alessandro Calefati  \and
        Muhammad Kamran Janjua \and 
        Ignazio Gallo 
}


\institute{S. Nawaz, A. Calefati, I. Gallo \at
              Department of Theoretical and Applied Science \\
              University of Insubria\\
              \email{\{snawaz,a.calefati,ignazio.gallo\}@uninsubria.it}           
           \and
           Muhammad Kamran Janjua \at
              School of Electrical Engineering and Computer Science \\
              National University of Sciences and Technology \\
              \email{mjanjua.bscs16seecs@seecs.edu.pk}
}

\date{Received: date / Accepted: date}

\maketitle

\begin{abstract}
The question we answer with this work is: ’can we convert a text document into an image to exploit best image classification models to classify documents?’
To answer this question we present a novel text classification method which converts a text document into an encoded image, using word embedding and capabilities of Convolutional Neural Networks (CNNs), successfully employed in image classification.
We evaluate our approach obtaining promising results on some well-known benchmark datasets for text classification.
This work allows the application of many of the advanced CNN architectures developed for Computer Vision to Natural Language Processing.
We test the proposed approach on a multi-modal dataset, proving that it is possible to use a single deep model to represent text and image in the same feature space.
\end{abstract}

\begin{figure}[!ht]
  \centering
  \includegraphics[width=0.6\columnwidth]{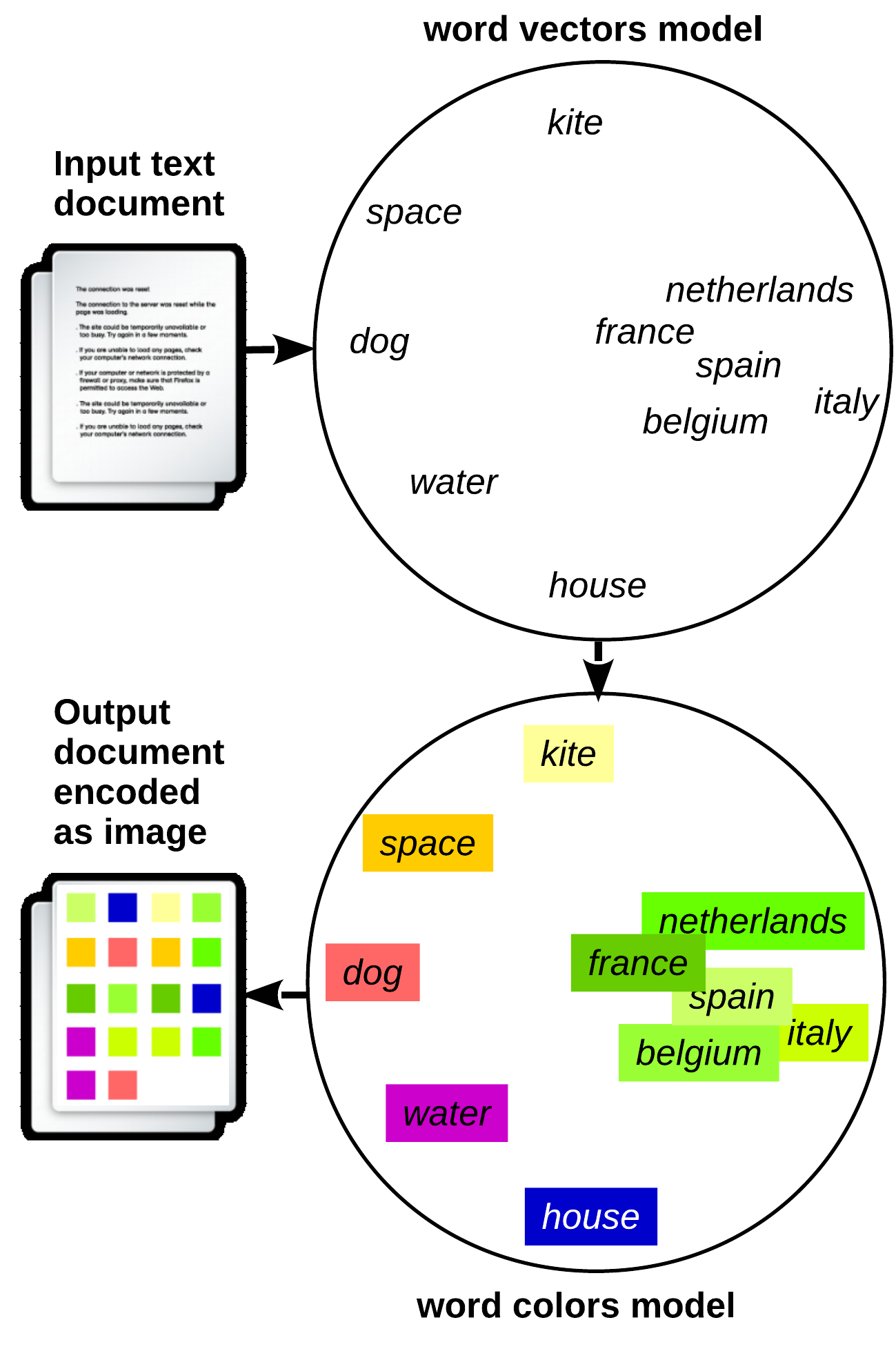}
  \caption{
  We exploited a well-known property of word embedding models: semantically correlated words obtain similar numerical representation.
  It turns out that if we interpret real valued vectors as set of colors, it is easy for a visual system to cope with relationships between words of a text document.
  It can be observed that green colored words are related to countries, while other words, are represented with different colors.
}
  \label{fig:idea}
\end{figure}

\begin{figure}
  \includegraphics[width=1.0\columnwidth]{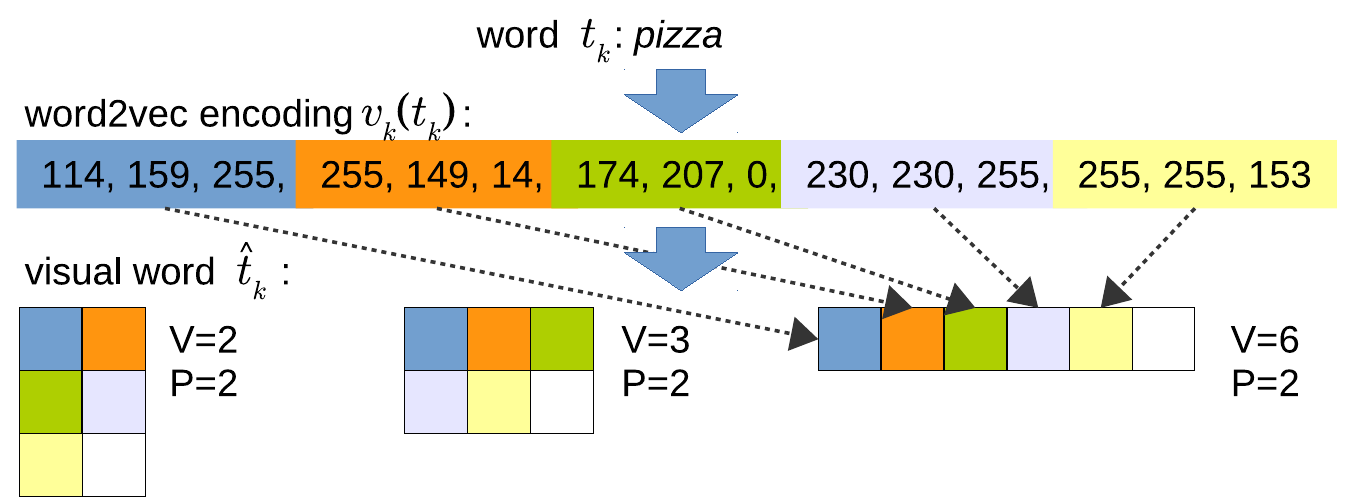}
     \caption{In this example, the word "\textit{pizza}" is encoded into a visual word $\hat{t}_k$ based on Word2Vec feature vector with length $15$. 
     This visual word can be transformed into different shapess, varying the V parameter (in this example $V = 2,3,6$ superpixels)}
  \label{fig:visual-word-example}
\end{figure}
\section{Introduction}

Text classification is a common task in Natural Language Processing.
Its goal is to assign a label to a text document from a predefined set of classes. 
Researchers have extensively focused on designing best features along with the choice of the best possible machine learning classifier. 
Traditional methods used in text classification such as n-grams~\cite{joachims1998text} captured the linguistic structure from a statistical point of view, however, recent methods successfully employ deep learning techniques and CNNs for text classification~\cite{kimconvolutional,zhang2015character}. 
Image classification has a similar aim to text classification and recently CNNs have become the de-facto standard in this field~\cite{krizhevsky2012imagenet,szegedy2015going}.

Word embedding models~\cite{Mikolov2013nips,pennington2014glove,le2014distributed} convert words or sentences into vectors of real numbers.
Typically these models are trained on large corpus of text documents to capture semantic relationships among words. 
Thus they can produce similar word embeddings for words occurring in similar contexts. 
We exploit this fundamental property of word embedding models to transform a text document into a sequence of colors, obtaining an artificial (encoded) image, as shown in Figure~\ref{fig:idea}.
Intuitively, semantically related words obtain similar colors or encodings in the encoded image while uncorrelated words are represented with different colors.

We present a novel text classification approach to cast text documents into visual domain.
Our approach transforms text documents into encoded images capitalizing on word embeddings. 
We evaluate the method on several large scale datasets obtaining promising and, in some cases, state-of-the-art results.
In summary, main contributions of our paper include:\\
-- We present an approach to transform text documents into encoded images based on Word2Vec word embedding.\\
-- A real-world multi-modal application based on the proposed approach that uses a single model to manage joint representations of image and encoded text.

A previous version of our encoding scheme was published in ICDAR 2017~\cite{gallosemantic2017}. In this work, we explore various parameters associated with encoding scheme. We extensively evaluate the improved encoding scheme on benchmark datasets. Furthermore, we present an application scenario based on our encoding scheme to fuse encoded text and image for multi-modal classification.

\begin{figure*}
  \centering
  \includegraphics[width=0.8\textwidth]{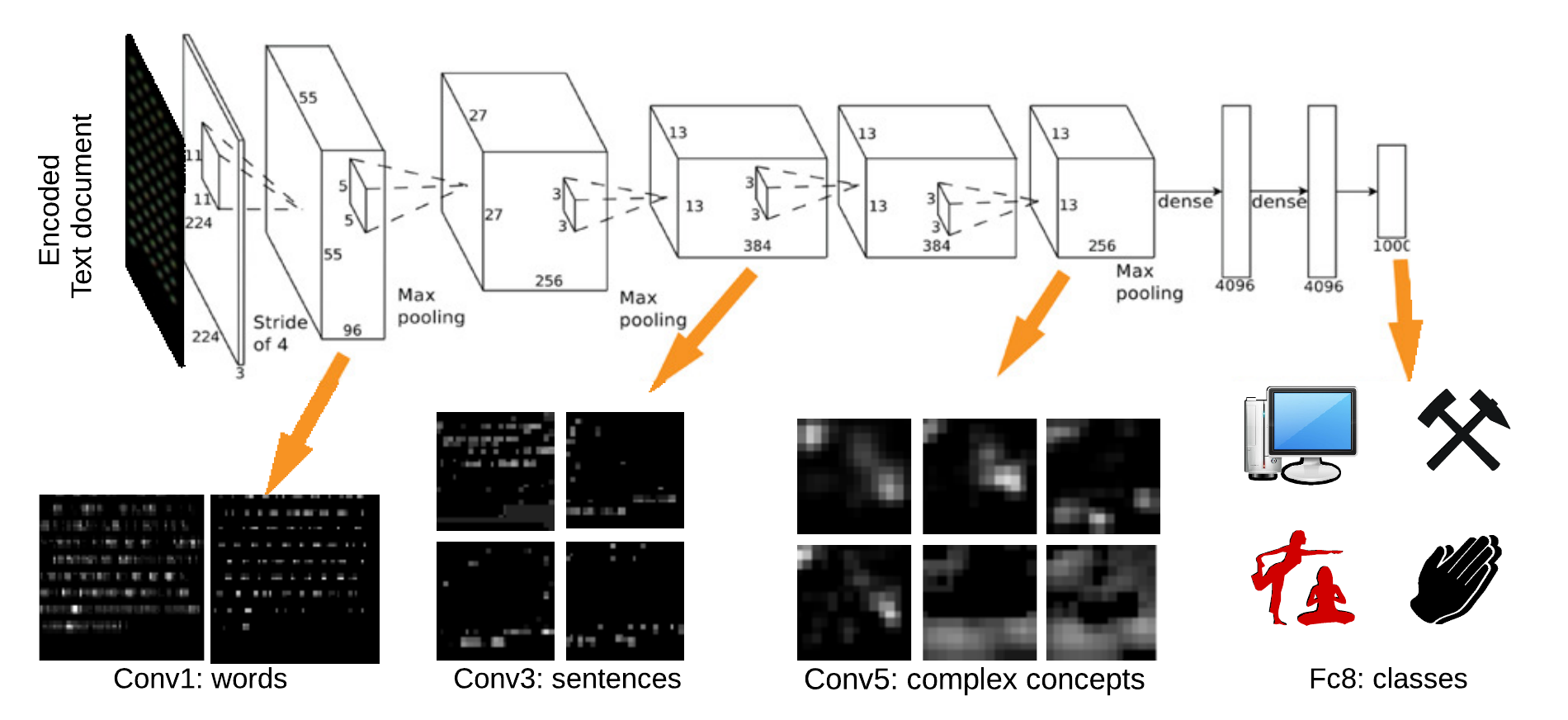}
  \caption{Starting from an encoded text document, the resulting image is classified by a CNN model normally employed in image classification.
  The first convolutional layers look some particular features of visual words while the remaining convolutional layers can recognize sentences and increasingly complex concepts.}
  \label{fig:proposed-model}
\end{figure*}

\section{Related Work}
\label{sec:related-work}

Much of the work with deep learning methods for text documents involved learning word vector representations through neural language models~\cite{Mikolov2013nips,pennington2014glove}.
This vector representation serves as a foundation of our work where word vectors are transformed into a sequence of colors. 
Tang et al. \cite{tang2014learning} adopted an ad-hoc model using three neural networks to encode the sentiment information from text. 
They concluded that word embedding models presented in~\cite{Mikolov2013nips} and~\cite{collobert2011natural} are not effective for sentiment classification.

\begin{figure}
  \centering
  \includegraphics[width=0.70\columnwidth]{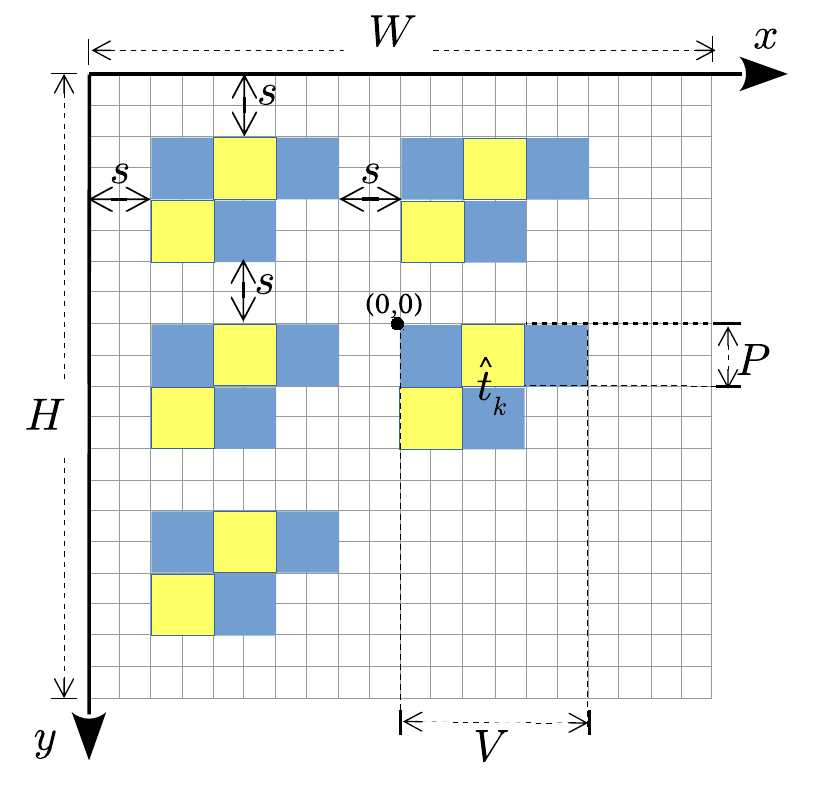}
  \caption{A graphical representation of our encoding scheme. 
  The figure shows an encoded image of size $W\times H$, with $5$ visual words $\hat{t}_k$ having width $V$ which contains superpixels of size $P\times P$.}
  \label{fig:visual-word}
\end{figure}

There are many works in literature like~\cite{zheng2017dual,wehrmann2018order,wang2016learning,wang2017learning} that deal with both images and texts to achieve semantic alignment.
However, they need to employ different deep models making their approaches computationally intensive, reinforced by~\cite{nawaz2018revisiting}.
Moreover, they have to create ad-hoc loss functions to achieve the result.
With our proposed method, to solve similar task we need just a single deep neural network.
Zhang et al.~\cite{zhang2015character} treated text as a raw signal at character level and employed deep architecture to perform text classification.
Similarly, Conneau et al.~\cite{conneau2017very} presented a deep architecture that operates at character level with $29$ convolutional layers to learn hierarchical representations of text.
This architecture is inspired by recent progress in computer vision~\cite{simonyan2014very,he2016deep}. 
In our work, we leverage on recent success in computer vision, but instead of adapting deep neural network to be fed with text information, we propose an approach that converts text documents into encoded text.
Once we have encoded text, we can apply state-of-the-art deep neural architectures.
We compared our proposed approach with deep learning models based on word embedding and lookup tables along with the method proposed in~\cite{zhang2015character} and~\cite{conneau2017very} using same datasets.
In Section~\ref{sec:experiments}, experimental results of our proposed approach are shown, highlighting that, in some cases, it overtakes state-of-the-art results while in other cases, it obtains comparative results.

\begin{figure*}
  \centering
  \begin{center}
  \begin{tabular}{ccccc}
  conv1 & conv2 & conv3 & conv4 & conv5\\
  \includegraphics[width=0.175\textwidth]{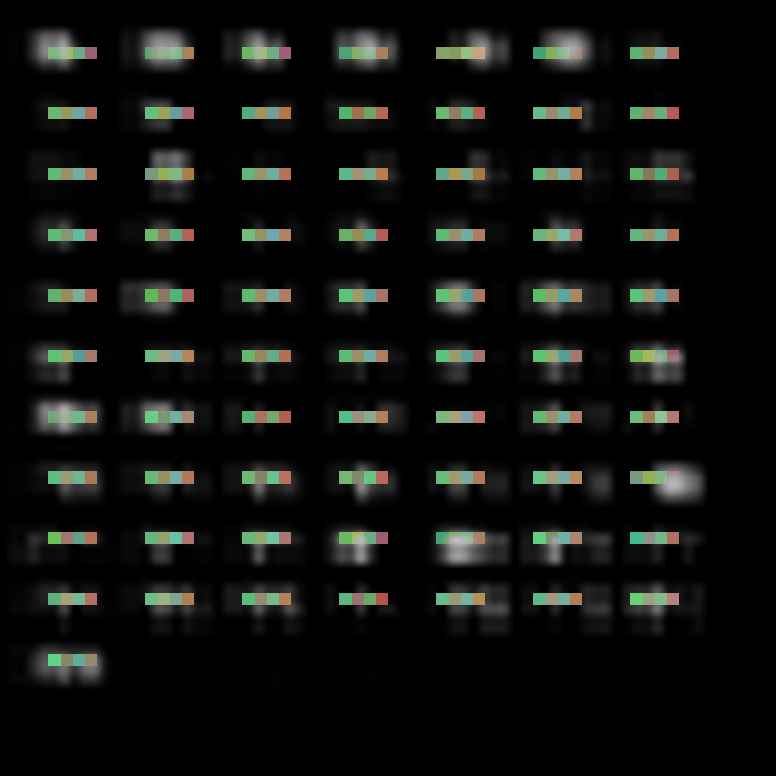} &
  \includegraphics[width=0.175\textwidth]{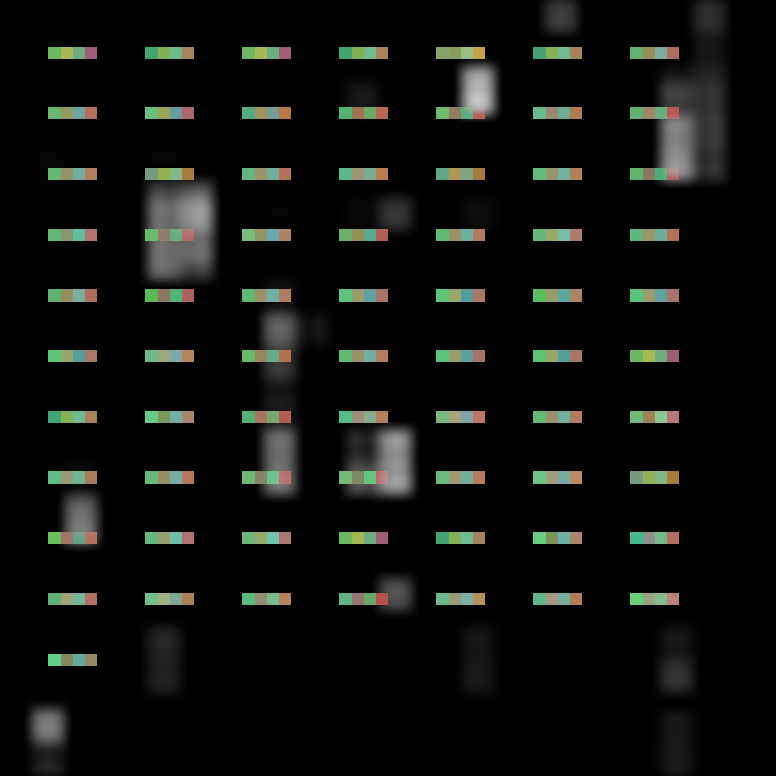} &
  \includegraphics[width=0.175\textwidth]{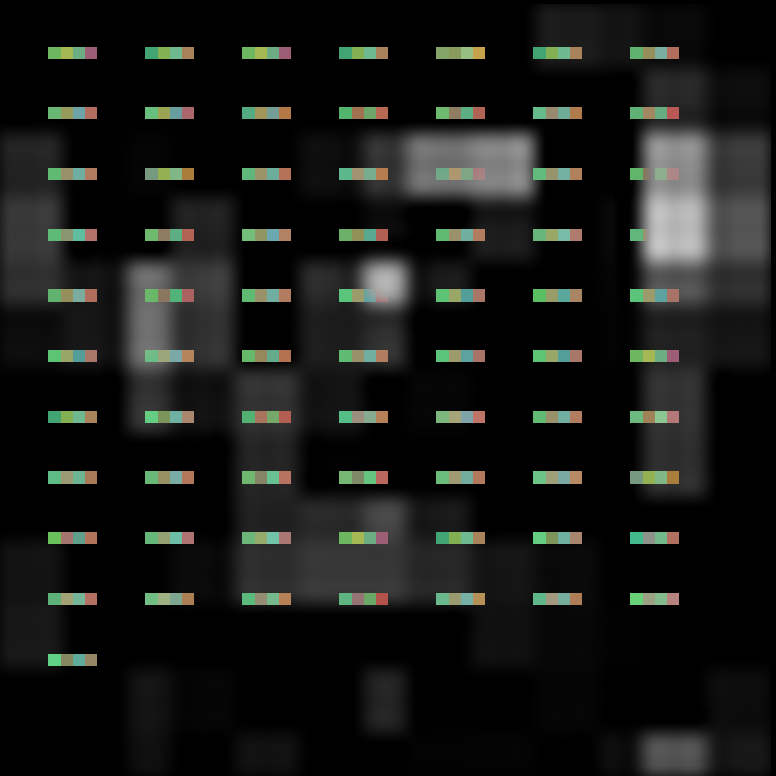} &
  \includegraphics[width=0.175\textwidth]{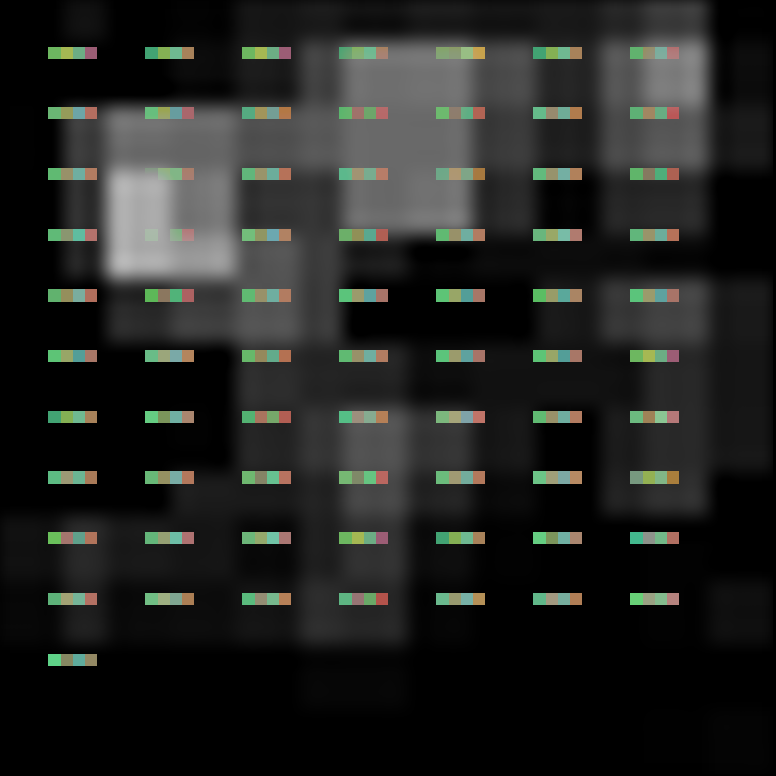} &
  \includegraphics[width=0.175\textwidth]{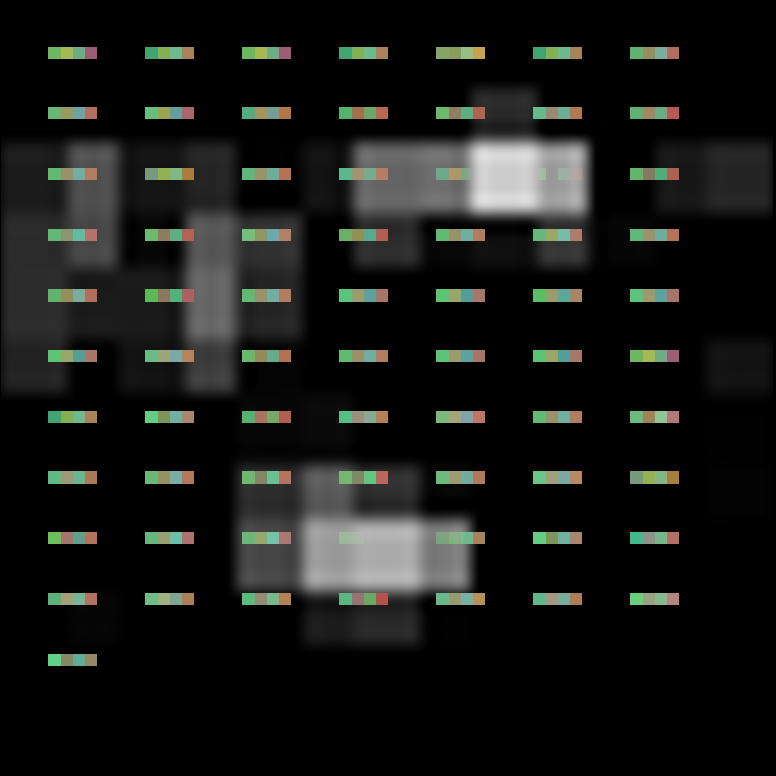} \\
  \end{tabular}
  \end{center}
  \caption{An example of five feature maps (conv1,\dots, conv5) displayed over the input image of the DBPedia dataset.
  In this particular example images are encoded with $12$ Word2Vec features and $16$ pixels of space between visual words.
  The convolutional map generated by the conv1 layer shows activations of individual superpixel or sequence of superpixels, while other convolutional layers show larger activation areas affecting more visual words.}
  \label{fig:feature-maps-example}
\end{figure*}

\section{Proposed Approach}
\label{sec:encodinbg}

\subsection{Encoding Scheme}
\label{sec:proposed-encoding-technique}

The proposed encoding approach is based on Word2Vec word embedding~\cite{Mikolov2013nips}.
We encode a word $t_k$ belonging to a document $D_i$ into an artificial image of size $W\times H$.
The approach uses a dictionary $F(t_k,v_k)$ with each word $t_k$ associated with a feature vector $v_k(t_k)$ obtained from a trained version of Word2Vec word embedding model.
Given a word $t_k$, we obtained a visual word $\hat{t}_k$ having width $V$ that contains subset of a feature vector, called superpixels (see example in Fig.~\ref{fig:visual-word-example}).
A superpixel is a square area of size $P\times P$ pixels with a uniform color that represents a sequence of contiguos features $(v_{k,j},v_{k,j+1},v_{k,j+2})$ extracted as a sub-vector of $v_k$.
A graphical representation is shown in Fig.~\ref{fig:visual-word}.
We normalize each component  $v_{k,j}$ to assume values in the interval $[0\dots255]$ with respect to $k$, then we interpret triplets from feature vector $v_k$ as RGB sequence.
For this very reason, we use feature vector with a length multiple of $3$.



The blank space $s$ around each visual word $\hat{t}_k$ plays an important role in the encoding approach.
We found out that the parameter $s$ is directly related to the shape of a visual word.
For example if $V = 16$ pixels then $s$ must also have a value close to $16$ pixels to let the network understand where a word ends and another begins.
A CNN in the first convolutional layer mainly extracts features for a single superpixel (see the \textit{conv1} example in Fig.~\ref{fig:feature-maps-example}) and so it is important to understand if two activated superpixels belong to same or different visual words.


Each convolutional layer produces convolutive maps from input to the last layer, as shown in Fig.~\ref{fig:proposed-model} and Fig.~\ref{fig:feature-maps-example}.
We noticed that the first convolutional layer recognizes some particular features of visual words associated to single or multiple superpixels.
Remaining CNN layers, instead, aggregate these simple activations to create increasingly complex relationships between words or parts of a sentence in a text document, similarly to image classification tasks.

\subsection{CNN Configuration}
\label{sec:cnn-configuration}

We encode the text document in an image to exploit the power of CNNs typically used in image classification. 
%
Usually, CNNs use ``\textit{mirror}'' data augmentation technique to obtain robust models in image classification.
In this work, this parameter is removed because mirroring an encoded text image changes the semantics of text documents and thus hampers the result.
%
Another way to increase the training data in image classification, is to ``\textit{crop}'' a certain number of patches from each training image and to use them as input to the CNN.
This process has been used and slightly improved results of our approach.

The ``\textit{stride}'' parameter is very primary in decreasing the complexity of the network, however, this value must not be bigger than the superpixel size, because larger values can skip too many pixels, which leads to information lost during the convolution, invalidating results.

\section{Dataset}
\label{sec:datasets}

In this section we provide an overview of several large-scale data sets introduced in~\cite{zhang2015character} which covers several classification tasks such as sentiment analysis, topic classification or news categorization. 
In addition, we used 20news-bydate dataset to test different parameters associated with the encoding approach. 
Table~\ref{tab:dataset-stats} reports a summary statistics of datasets.

\textbf{20news-bydate.} We used 20news-bydate version of 20news dataset available on the web\footnote{\small http://qwone.com/$\sim$jason/20Newsgroups/}. 
We selected $4$ major categories: comp, politics, rec and religion as described in~\cite{hingmire2013document}. 
These categories contain $7,975$ and $5,319$ samples for training and test sets respectively.

\textbf{AG's news corpus.} We used the version created by Zhang~\textit{et al.} who selected $4$ largest classes from AG news corpus on the web~\cite{zhang2015character}\footnote{\small http://www.di.unipi.it/$\sim$gulli/AG\_corpus\_of\_news\_articles.html}. 
The number of training instances for each class is $30,000$ and $1,900$ for testing with each instance containing class index, title and description fields. 

\textbf{Sogou news corpus.} Zhang~\textit{et al.} proposed a revised version of this dataset by selecting top $5$ categories from the SogouCA and SogouCS news corpora. 
The number of training instances selected for each class is $90,000$ and $12,000$ for testing with each instance containing title and content fields~\cite{zhang2015character}. 
This dataset is in Chinese language, Zhang~\textit{et al.} used a pypinyin package combined with the jieba Chinese segmentation system to produce the pinyin, a phonetic romanization of Chinese characters of words so that they were able to employ the same models for the English language without changes~\cite{zhang2015character}.

\textbf{DBPedia ontology dataset.} This dataset was created by selecting $14$ non-overlapping classes from DBPedia 2014. 
$40,000$ training samples and $5,000$ testing samples were randomly selected from these classes with each instance containing the title and abstract of each Wikipedia article.

\textbf{Yelp reviews.} Zhang~\textit{et al.} used Yelp Dataset Challenge 2015 dataset to perform two classification tasks~\cite{zhang2015character}. 
The first task predicts stars assigned by the user and the other task predicts the polarity label by considering stars 1 and 2 as negative while 3 and 4 as positive. 
These two tasks led to the construction of two datasets: the full dataset has $130,000$ training samples and $10,000$ testing samples for each star while the polarity dataset has $280,000$ training samples and $19,000$ test samples for each polarity.

\textbf{Yahoo! Answers dataset.} Zhang~\textit{et al.} obtained Yahoo! Answers Comprehensive Questions and Answers version 1.0 dataset through the Yahoo! Webscope program~\cite{zhang2015character}.  
The $10$ largest classes are selected from this corpus to create a topic classification dataset with each instance containing question title, question content and best answer. 
Each class contains $140,000$ training and $5,000$ testing samples.  

\textbf{Amazon reviews.} Zhang~\textit{et al.} used Amazon review dataset from Stanford Network Analysis Project (SNAP) to perform full score prediction and create polarity prediction datasets~\cite{zhang2015character}. 
The full dataset contains $600,000$ training samples and $130,000$ testing samples in each class, whereas the polarity dataset contains about $1,800,000$ training and $200,000$ testing samples. 

\begin{table}[htbp]
\caption{ Statistics of 20news-bydate and large-scale datasets presented in Zhang~\textit{et al.}~\cite{zhang2015character}.
The $2$ rightmost columns show the average (Avg) and standard deviation (Std) for the number of words contained in text documents.}
\label{tab:dataset-stats}
\begin{minipage}{.001\linewidth}
\scalebox{0.80}{
\begin{tabular}{lccccc}
\hline
\textbf{Dataset}       & {\textbf{Classes}} & \textbf{Training}  &  \textbf{Test}  &  \textbf{Avg}   &  \textbf{Std}  \\ \hline
20news-bydate          & 4                  & 7,977                  & 7,321    & 339 & 853 \\
AG's News              & 4                  & 120,000                & 7,600    & 43  & 13  \\ 
Sogou News             & 5                  & 450,000                & 60,000   & 47  & 73  \\ 
DBPedia                & 14                 & 560,000                & 70,000   & 54  & 25  \\
Yelp Review Polarity   & 2                  & 560,000                & 38,000   & 138 & 128 \\ 
Yelp Review Full       & 5                  & 650,000                & 50,000   & 140 & 127 \\ 
Yahoo! Answers         & 10                 & 1,400,000              & 60,000   & 97  & 105 \\ 
Amazon Review Full     & 5                  & 3,000,000              & 650,000  & 91  & 49  \\ 
Amazon Review Polarity & 2                  & 3,600,000              & 400,000  & 90  & 49  \\ 
\hline 
\end{tabular}}
\end{minipage}
\end{table}

\begin{figure}
  \centering
  \includegraphics[width=1.0\columnwidth]{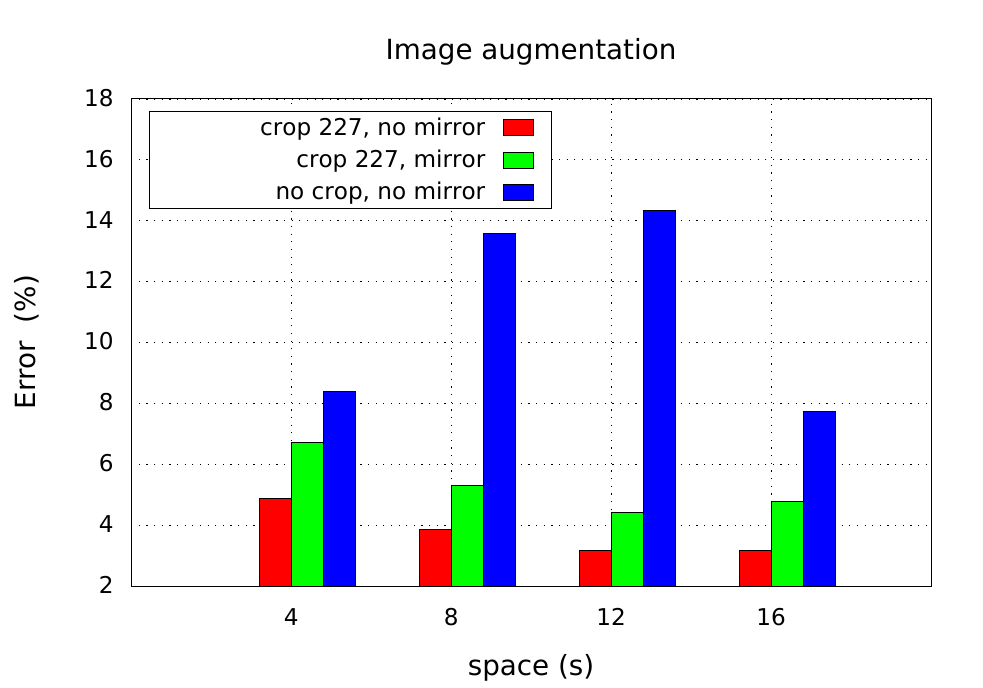}
  \caption{Classification error using data augmentation: (\textit{mirror} and \textit{crop}) over the 20news-bydate test set. 
  }
  \label{fig:image-augmentations-results}
\end{figure}

\begin{figure*}[t]
  \centering
  \includegraphics[width=0.17\textwidth]{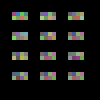}
  \includegraphics[width=0.17\textwidth]{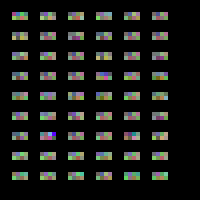}
  \includegraphics[width=0.17\textwidth]{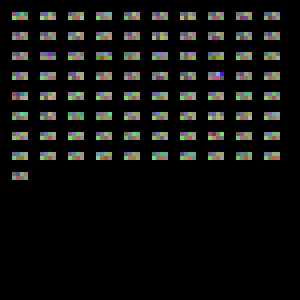}
  \includegraphics[width=0.17\textwidth]{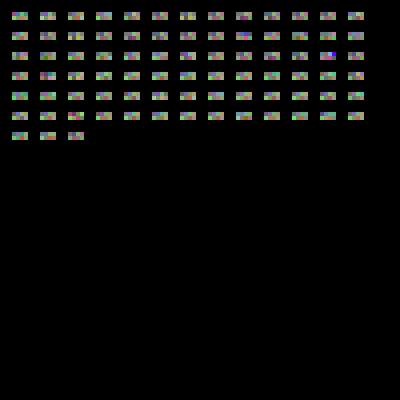}
  \includegraphics[width=0.17\textwidth]{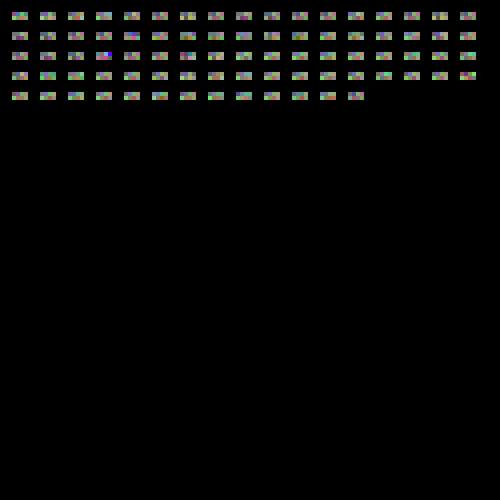}
  \caption{Five different sizes of encoded image  ($100\times100$, $200\times200$, $300\times300$, $400\times400$, $500\times500$) obtained using the same document belonging to the 20news-bydate dataset. 
  All images use the same encoding with $24$ Word2Vec features, space $s=12$, superpixel size $4\times4$.
  It is important to note that the two leftmost images cannot represent all words in the document due to the small size.}
  \label{fig:image-size-configuration}
\end{figure*}

\begin{figure*}
  \centering
  \includegraphics[width=0.17\textwidth]{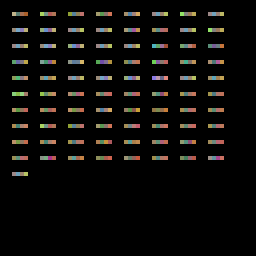}
  \includegraphics[width=0.17\textwidth]{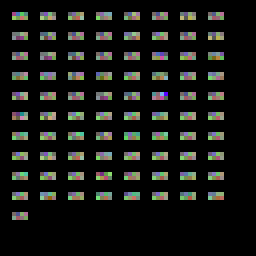}
  \includegraphics[width=0.17\textwidth]{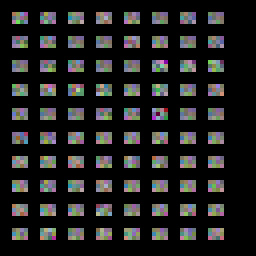}
  \includegraphics[width=0.17\textwidth]{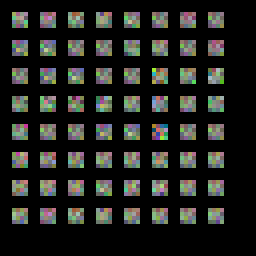}
  \includegraphics[width=0.17\textwidth]{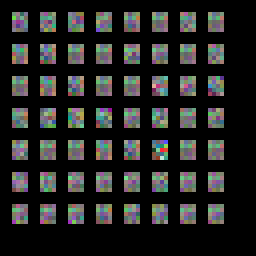}
  \caption{Five encoded images obtained using different Word2Vec features length and using the same document belonging to the 20news-bydate dataset.
  All the images are encoded using space $s=12$, superpixel size $4\times4$, image size $= 256\times256$ and visual word width $V=16$.
  The two leftmost images contain all words in the document encoded with $12$ and $24$ Word2Vec features respectively, while $3$ rightmost encoded images with $36$, $48$ and $60$ features length cannot encode entire documents.}
  \label{fig:image-size-example}
\end{figure*}

\begin{table*}[tpb]
\caption{Comparison of different parameters over the 20news-bydate dataset.
    In the leftmost table we changed the size of the encoded image from $100\times100$ to $500\times500$ and the crop size is also changed by multiplying the image size with a constant i.e. $1.13$.
    Here \textit{sp} stands for superpixel, \textit{w2v} is for number of Word2Vec features, \textit{Mw} stands for Max number of visual words that an image can contain and \textit{\#w} is the number of text documents in the test set having a greater number of words than \textit{Mw}.
    We fixed the remaining non-specified parameters as follow: $s=12$, $V=4$, $sp=4$, image size$=256$.}
\label{tab:comp-image-sizes}
    \begin{minipage}{.31\linewidth}
      \centering
      \begin{tabular}{crr}
      \hline
        \textbf{image size} & \textbf{crop}  & \textbf{error} \\ \hline
        $500\times500$ & 443 & \textbf{8.63} \\ 
        400x400 & 354 & 9.30 \\ 
        300x300 & 266 & 10.12 \\ 
        200x200 & 177 & 10.46 \\ 
        100x100 & 88 & 15.70 \\  \hline
      \end{tabular}
    \end{minipage}
    \begin{minipage}{.17\linewidth}
      \centering
        \begin{tabular}{cr}
        \hline
          \textbf{sp} & \textbf{error} \\ \hline
          5x5 & 8.96 \\ 
          4x4 & \textbf{8.87} \\ 
          3x3 & 10.27 \\ 
          2x2 & 10.82 \\ 
          1x1 & 10.89 \\ \hline
        \end{tabular}
    \end{minipage}
    \begin{minipage}{.185\linewidth}
      \centering
        \begin{tabular}{cr}
        \hline
        \textbf{stride} & \textbf{error} \\ \hline
        5 & 8.7 \\ 
        4 & 8.87 \\ 
        3 & 8.33 \\ 
        2 & \textbf{7.78} \\ 
        1 & 12.5 \\ \hline
        \end{tabular}
    \end{minipage}
    \begin{minipage}{.18\linewidth}
      \centering
        \begin{tabular}{crrr}
        \hline
        \textbf{w2v} & \textbf{Mw} & \textbf{\#w} & \textbf{error} \\ \hline
        12 & 180 & 50\% & 9.32 \\ 
        24 & 140 & 64\% & 8.87 \\ 
        36 & 120 & 71\% & \textbf{7.20} \\ 
        48 & 100 & 79\% & 8.21 \\ 
        60 & 90  & 83\% & 20.66 \\ \hline
        \end{tabular}
    \end{minipage} 
\end{table*}

\begin{figure*}[t]
 \centering
  \begin{minipage}{.25\linewidth}
\begin{tabular}{ >{\centering\arraybackslash}m{1.5cm}  >{\centering\arraybackslash}m{2.5cm} }
  $VW-1$ & \includegraphics[width=0.26\textwidth]{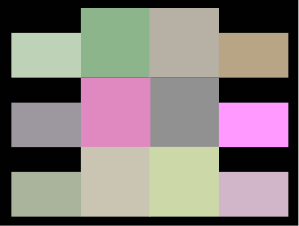} \\
  $VW-2$ & \includegraphics[width=0.26\textwidth]{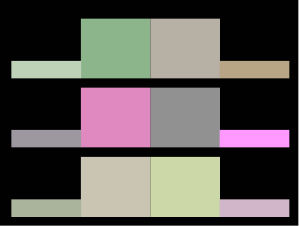}  \\
  $VW-3$ & \includegraphics[width=0.40\textwidth]{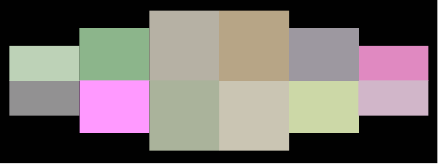} \\
  $VW-4$ & \includegraphics[width=0.40\textwidth]{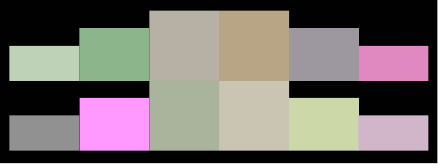} \\
  $VW-5$ & \includegraphics[width=0.40\textwidth]{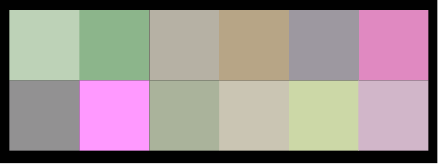} \\
    \end{tabular}
  \end{minipage} 
  \begin{minipage}{.5\linewidth}
    \centering
    \includegraphics[width=0.9\textwidth]{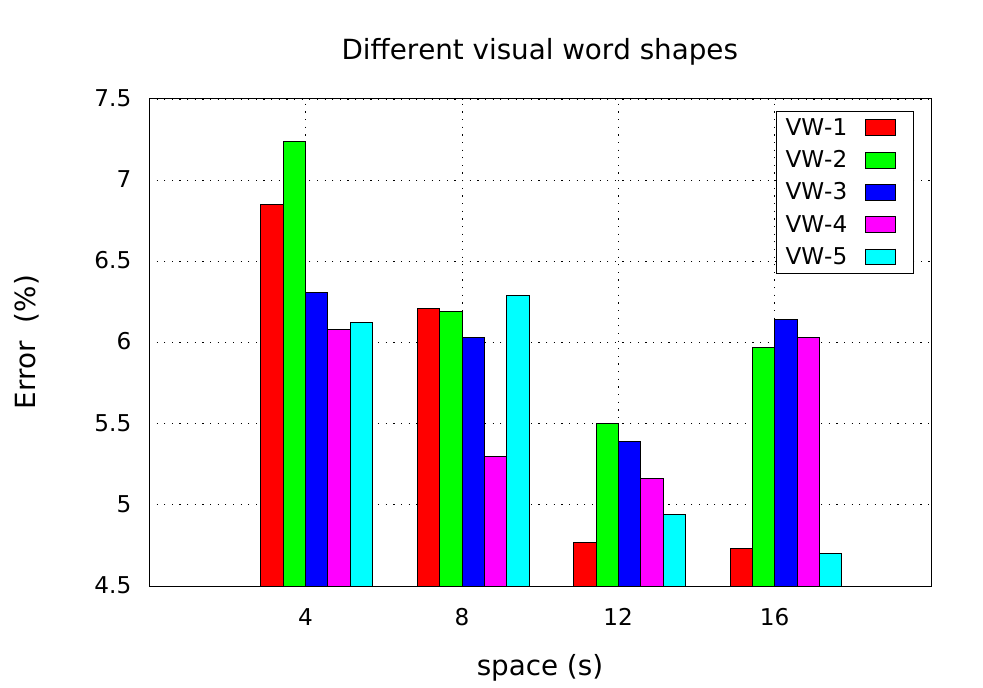}
  \end{minipage} 
  \caption{On the left, five different designs for visual words ($VW$) represented by $36$ Word2Vec features, over the 20news-bydate dataset.
  The width V of these words is $4$ for the first two on the top and $6$ for the rest.
  The first four visual words consist of super pixels with different shapes to form particular visual words.
  On the right, a comparison over these different shapes of visual words.}
  \label{fig:different-visual-words-comparison}
\end{figure*}

\begin{table}
\centering
\caption{Comparison between CNNs trained with different configurations on our proposed approach.
The width $V$ (in superpixels) of visual words is fixed while the Word2Vec encoding vector size and space $s$ (in pixel) varies.
$H$ is the height of visual word obtained.}
\label{tab:height-space}
\begin{tabular}{lcccc}
\hline
$s$ & $V$ & $H$ & w2v feat.& error (\%) \\ \hline
4  & 4 & 1  & 12 &  7.63 \\ 
8  & 4 & 1  & 12 &  5.93 \\ 
12 & 4 & 1  & 12 &  \textbf{4.45} \\ 
16 & 4 & 1  & 12 &  4.83 \\ \hline
4  & 4 & 2  & 24 &  6.94 \\ 
8  & 4 & 2  & 24 &  5.60  \\ 
12 & 4 & 2  & 24 &  5.15 \\ 
16 & 4 & 2  & 24 &  \textbf{4.75} \\ \hline
4  & 4 & 3  & 36 &  6.72 \\ 
8  & 4 & 3  & 36 &  5.30  \\ 
12 & 4 & 3  & 36 &  \textbf{4.40}  \\ 
16 & 4 & 3  & 36 &  4.77 \\ \hline
\end{tabular}
\end{table}

\section{Experiments}
\label{sec:experiments}

The aim of these experiments is threefold: 
(i) to evaluate configuration parameters associated with the encoding approach; 
(ii) to compare the proposed approach with other deep learning methods;
(iii) to validate the proposed approach on a real-world application scenario.

In our experiments, percentage error is used to measure the classification performance.
The encoding approach mentioned in Section~\ref{sec:proposed-encoding-technique} produces encoded image based on Word2Vec word embedding. 
These encoded images can be used to train and test a CNN.
We used AlexNet~\cite{krizhevsky2012imagenet} and Googlenet~\cite{szegedy2015going} architectures as base models.
We used a publicly available Word2Vec word embedding with default configuration parameters as in~\cite{Mikolov2013nips} to train word vectors on all datasets.
Normally, Word2Vec is trained on a large corpus and used in different contexts. 
However, in our work, we trained this model with the same training set for each dataset.

\subsection{Parameters setting}
We used 20news-bydate dataset to perform a series of experiments with different settings to find out the best configuration for the proposed method.
In our first experiment, we changed the space $s$ among visual words and the number of Word2Vec features to identify relationships between these parameters.
The result is shown in Table~\ref{tab:height-space}; we used the best result from this table to perform other experiments shown in Table~\ref{tab:result-comparison}.
We obtained a lower percentage error with higher values of $s$ parameter and higher number of Word2Vec features.
The length of feature vector $v_k(t_k)$ depends on the nature of the dataset. 
For example in Fig.~\ref{fig:image-size-example}, a text document composed of a large number of words cannot be encoded completely using high number of Word2Vec features, because each visual word occupies more space in the encoded image.
Furthermore, we found out that error does not decrease linearly with the increase of Word2Vec features, as shown in Table~\ref{tab:comp-image-sizes}.

We tested various shapes for visual words before selecting the best one, as shown in Fig.~\ref{fig:different-visual-words-comparison} (on the left). 
We showed that with rectangular shaped visual words, we obtained higher results, as highlighted in Fig.~\ref{fig:different-visual-words-comparison} (on the right).
Moreover, the space $s$ between visual words plays an important role in the final classification, in fact using a high value for the $s$ parameter, the convolutional layer can effectively distinguish among visual words, also demonstrated from the results in Table~\ref{tab:height-space}.
As shown in Fig.~\ref{fig:feature-maps-example}, the first level of a CNN (\textit{conv1}) specializes convolution filters in the recognition of a single superpixel. 
Hence, it is important to distinguish between superpixels of different visual words by increasing the parameter $s$.

These experiments led us to the conclusion that we have a trade-off between the number of Word2Vec features to encode each word and the number of words that can be represented in an image.
In fact, increasing the number of Word2Vec features increases the space required in the encoded image to represent a single word.
Moreover, this aspect affects the maximum number of words that can fit in an image.
The choice of this parameter must be done considering the nature of the dataset, whether it is characterized by short or long text documents.
For our experiments, we used a value of $36$ for Word2Vec features, considering results presented in Table~\ref{tab:comp-image-sizes}.

\subsection{Data augmentation}
We showed that the \textit{mirror} data augmentation technique, successfully used in image classification, is not recommended here because it changes the semantics of the encoded words and can deteriorate the classification performance.
Results are presented in Fig.~\ref{fig:image-augmentations-results}. 
In addition, we showed that increasing the number of training samples by using the \textit{crop} parameter, results are improved.
More precisely, during the training and test phases, $10$ random $227\times227$ crops are extracted from a $256\times256$ image (or proportional crop for different image size, 
as reported in the leftmost Table~\ref{tab:result-comparison}) and then fed to the network.
During the testing phase we extracted a $227\times227$ patch from the center of the image.

\begin{table*}
\centering
\caption{Testing error of our encoding approach on $8$ datasets with Alexnet and GoogleNet. 
In addition, results obtained by Zhang~\textit{et al.}~\cite{zhang2015character} and Conneau~\textit{et al.}~\cite{conneau2017very} are included.
}
\label{tab:result-comparison}
\begin{tabular}{lcccccccc}
\hline
\textbf{Model} & {\textbf{AG}} & \textbf{Sogou} & \textbf{DBP.} & \textbf{Yelp P.} & \textbf{Yelp F.} & \textbf{Yah. A.} & \textbf{Amz. F.} & \textbf{Amz. P.} \\ \hline
Zhang \textit{et al.} & \textbf{7.64} & \textbf{2.81} & 1.31 & 4.36 & 37.95 & 28.80 & 40.43 & 4.93                 \\ 
Conneau \textit{et al.} & 8.67 & 3.18 & 1.29 & \textbf{4.28} & \textbf{35.28} & 26.57 & \textbf{37.00} & 4.28 \\ \hline
Encoding scheme + AlexNet & 9.19 & 8.02 & 1.36 & 11.55 & 49.00 & 25.00 & 43.75 & 3.12  \\  
Encoding scheme + GoogleNet & 7.98 & 6.12 & \textbf{1.07} & 9.55 & 43.55 & \textbf{24.90} & 40.35 & \textbf{3.01}          \\  \hline 
\end{tabular}
\end{table*}

\subsection{Encoded image size}
We used various image sizes for the encoding approach.
Fig.~\ref{fig:image-size-configuration} shows artificial images built on top of Word2Vec features with different sizes. 
As illustrated in Table~\ref{tab:comp-image-sizes}, percentage error decreases by increasing the size of an encoded image; however, we observed that sizes above $300\times300$ is computationally intensive; hence, this lead us to chose an image size of $256\times256$, typically used in AlexNet and GoogleNet architectures.

\subsection{Comparison with other state-of-the-art text classification methods}
We compared our approach with state-of-the-art methods.
Zhang \textit{et al.} presented a detailed analysis between traditional and deep learning methods.
From the paper, we selected best results and reported them in Table~\ref{tab:result-comparison}.
In addition, we also compared our results with Conneau \textit{et al.}
We obtained state-of-the-art results on DBPedia, Yahoo Answers! and Amazon Polarity datasets, while comparative results on AGnews, Amazon Full  and Yelp Full datasets.
However, we obtained higher error on Sogou dataset due to the translation process explained in  Section~\ref{sec:datasets}.



\subsection{Comparison with state-of-the-art CNNs}
We obtained better performance using GoogleNet, as expected.
This lead us to believe that recent state-of-the-art network architectures, such as Residual Network would further improve results. 
To work successfully with huge datasets and powerful models, a high-end hardware and large training time are required, thus we conducted experiments only on 20news-bydate dataset with three network architectures: AlexNet, GoogleNet and ResNet.
Results are shown in Table~\ref{tab:result-comparison-netrworks}. 
We achieved better performance with powerful network architecture. 

\begin{table}
\centering
\caption{Percentage errors on 20news-bydate dataset with three different CNNs.
}
\label{tab:result-comparison-netrworks}
\begin{tabular}{lc}
\hline
CNN architecture          		   &  error   \\ \hline
Encoding scheme + AlexNet          &  4.10    \\ 
Encoding scheme + GoogleNet        &  3.81    \\ 
Encoding scheme + ResNet           &  2.95    \\ \hline
\end{tabular}
\end{table}

\setlength{\tabcolsep}{4pt}
\begin{table}
\begin{center}
\caption{Percentage error between proposed approach and single sources.}
\label{tab:main-results}
\begin{tabular}{lccc}
\hline
Dataset & Image & Text & \textbf{Proposed}\\
\hline
Amazon      	 		& 53.93		& 35.93 	& \textbf{27.48}\\
Ferramenta  	 		& 7.64		& 12.14 	& \textbf{5.16}\\

\hline
\end{tabular}
\end{center}
\end{table}
\setlength{\tabcolsep}{1.4pt}

\begin{figure}
  \centering 
  \begin{tabular}{ccc}
  \frame{\includegraphics[width=0.12\textwidth]{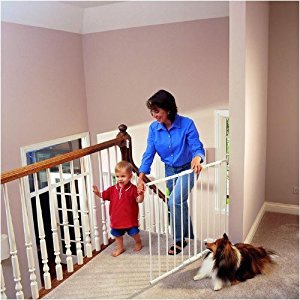}} &
  \frame{\includegraphics[width=0.12\textwidth]{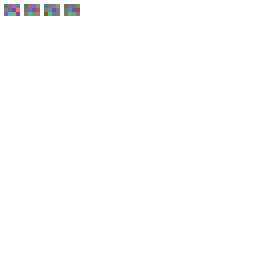}} &
  \frame{\includegraphics[width=0.12\textwidth]{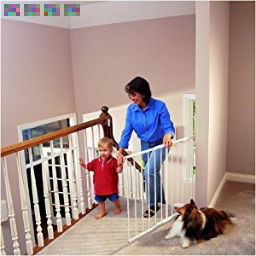}}  \\
  (a) & (b) & (c) \\
  \end{tabular}
  \caption{An example of multi-modal fusion from the Amazon dataset belonging to the class "Baby". 
  (a) shows the original image, (b) is a encoded text image and (c) shows the image with the superimposition of the encoded text in the upper part. The text in this example contains only the following 4 words "Kidco Safeway white G2000". 
 The size of all images is $256\times256$.
  }
  \label{fig:multimodal-fusion-example}
\end{figure}

\subsection{An Application Scenario}
\subsubsection{Classification Scenario}
One of the main advantages of the proposed approach is the ability of converting a text document into the same feature space of an image.
We therefore believe that the proposed approach can be exploited to improve the solution of problems requiring the combined use of image and text.
For example, an advertisement on e-commerce websites consists of image and text description, therefore it becomes useful to exploit both these sources in a multi-modal strategy.

Therefore, we use two multi-modal datasets to demonstrate that our approach brings significant benefits.
The first dataset named Ferramenta dataset~\cite{gallo2017multimodal} consists of $88,010$ advertisements split in $66,141$ adverts for train and $21,869$ for test sets, belonging to $52$ classes.
We used another publicly available real-world dataset called \textit{Amazon Product Data}~\cite{he2016ups}.
We randomly selected $10,000$ adverts belonging to $22$ classes.
Finally, we split $10,000$ advertisements for each class into train and test sets with $7,500$ and $5,000$ advertisements respectively.

Both datasets provide a text and representative image for each advertisement.
Text descriptions in Ferramenta dataset are in Italian language and we preferred not to translate to English, because we believe that the translation process would alter the nature of the text descriptions.
However, text descriptions in Amazon Product Data dataset are in English.
We compare the classification of advertisement using only the encoded text description, only the image and their combination.
A sample image for these experiments is shown in Fig.~\ref{fig:multimodal-fusion-example}.
The model trained on images only for Amazon dataset obtained the following first two predictions: \textbf{$\textbf{77.42\%}$ Baby}  and \textbf{$\textbf{11.16\%}$ Home and Kitchen} on this example. 
While the model trained on the multi-modal Amazon dataset obtained the following first two predictions: \textbf{$\textbf{100\%}$ Baby} and $\textbf{0\%}$ \textbf{Patio Lawn and Garden} for the same example.
This indicates that our encoding technique leads to a better classification result than the classification of text or image alone.
Table~\ref{tab:main-results} shows that the combination of text and image into a single image, outperforms best result obtained using only a uni-modality on both multi-modal datasets.

\subsubsection{Retrieval Scenario}
The work in~\cite{nawaz2018revisiting} employs our encoding scheme for cross-modal retrieval system. With our encoding scheme, the retrieval system uses a single network trained in an end-to-end fashion.
The results presented verify that raw text features are not necessary to map text descriptions to a similar embedding space with their respective images.

\section{Conclusion}
\label{sec:con}
In this work, we presented a novel text classification approach to transform Word2Vec features of text documents into artificial images to exploit CNNs capabilities for text classification.
We obtained state-of-the-art results on some datasets while in other cases our approach obtained comparative results. 
We showed that the CNN model generally used for image classification is successfully employed for different task such as text classification. 
As shown in the experiment section, the trend in results clearly show that, we can further improve results with more recent and powerful deep learning models for image classification. 

\section{Acknowledgment}
We gratefully acknowledge the support of NVIDIA Corporation, which donated the GeForce GTX 980 GPU used in some experiments.
We also acknowledge E4 Computer Engineering S.p.a. for providing an NVIDIA DGX-1, which allowed us to run experiments on computationally intensive datasets.

\bibliographystyle{spmpsci}
\bibliography{refs}

\begin{thebibliography}{10}
\providecommand{\url}[1]{{#1}}
\providecommand{\urlprefix}{URL }
\expandafter\ifx\csname urlstyle\endcsname\relax
  \providecommand{\doi}[1]{DOI~\discretionary{}{}{}#1}\else
  \providecommand{\doi}{DOI~\discretionary{}{}{}\begingroup
  \urlstyle{rm}\Url}\fi

\bibitem{collobert2011natural}
Collobert, R., Weston, J., Bottou, L., Karlen, M., Kavukcuoglu, K., Kuksa, P.:
  Natural language processing (almost) from scratch.
\newblock Journal of Machine Learning Research \textbf{12}(Aug), 2493--2537
  (2011)

\bibitem{conneau2017very}
Conneau, A., Schwenk, H., Barrault, L., Lecun, Y.: Very deep convolutional
  networks for text classification.
\newblock In: Proceedings of the 15th Conference of the European Chapter of the
  Association for Computational Linguistics: Volume 1, Long Papers, vol.~1, pp.
  1107--1116 (2017)

\bibitem{gallo2017multimodal}
Gallo, I., Calefati, A., Nawaz, S.: Multimodal classification fusion in
  real-world scenarios.
\newblock In: Document Analysis and Recognition (ICDAR), 2017 14th IAPR
  International Conference on, vol.~5, pp. 36--41. IEEE (2017)

\bibitem{he2016deep}
He, K., Zhang, X., Ren, S., Sun, J.: Deep residual learning for image
  recognition.
\newblock In: Proceedings of the IEEE conference on computer vision and pattern
  recognition, pp. 770--778 (2016)

\bibitem{he2016ups}
He, R., McAuley, J.: Ups and downs: Modeling the visual evolution of fashion
  trends with one-class collaborative filtering.
\newblock In: proceedings of the 25th international conference on world wide
  web, pp. 507--517. International World Wide Web Conferences Steering
  Committee (2016)

\bibitem{hingmire2013document}
Hingmire, S., Chougule, S., Palshikar, G.K., Chakraborti, S.: Document
  classification by topic labeling.
\newblock In: Proceedings of the 36th international ACM SIGIR conference on
  Research and development in information retrieval, pp. 877--880. ACM (2013)

\bibitem{joachims1998text}
Joachims, T.: Text categorization with support vector machines: Learning with
  many relevant features.
\newblock Machine learning: ECML-98 pp. 137--142 (1998)

\bibitem{kimconvolutional}
Kim, Y.: Convolutional neural networks for sentence classification.
\newblock In: Proceedings of the 2014 Conference on Empirical Methods in
  Natural Language Processing (EMNLP), pp. 1746--1751. Association for
  Computational Linguistics (2014)

\bibitem{krizhevsky2012imagenet}
Krizhevsky, A., Sutskever, I., Hinton, G.E.: Imagenet classification with deep
  convolutional neural networks.
\newblock In: Advances in neural information processing systems, pp. 1097--1105
  (2012)

\bibitem{Mikolov2013nips}
Mikolov, T., Sutskever, I., Chen, K., Corrado, G.S., Dean, J.: Distributed
  representations of words and phrases and their compositionality.
\newblock In: C.J.C. Burges, L.~Bottou, M.~Welling, Z.~Ghahramani, K.Q.
  Weinberger (eds.) Proceedings of the 26th International Conference on Neural
  Information Processing Systems, NIPS'13, pp. 3111--3119. Curran Associates,
  Inc. (2013)

\bibitem{simonyan2014very}
Simonyan, K., Zisserman, A.: Very deep convolutional networks for large-scale
  image recognition.
\newblock arXiv preprint arXiv:1409.1556  (2014)

\bibitem{szegedy2015going}
Szegedy, C., Liu, W., Jia, Y., Sermanet, P., Reed, S., Anguelov, D., Erhan, D.,
  Vanhoucke, V., Rabinovich, A.: Going deeper with convolutions.
\newblock In: Proceedings of the IEEE conference on computer vision and pattern
  recognition, pp. 1--9 (2015)

\bibitem{tang2014learning}
Tang, D., Wei, F., Yang, N., Zhou, M., Liu, T., Qin, B.: Learning
  sentiment-specific word embedding for twitter sentiment classification.
\newblock In: ACL (1), pp. 1555--1565 (2014)

\bibitem{wehrmann2018order}
Wehrmann, J., Mattjie, A., Barros, R.C.: Order embeddings and character-level
  convolutions for multimodal alignment.
\newblock Pattern Recognition Letters \textbf{102}, 15--22 (2018)

\bibitem{zhang2015character}
Zhang, X., Zhao, J., LeCun, Y.: Character-level convolutional networks for text
  classification.
\newblock In: Advances in neural information processing systems, pp. 649--657
  (2015)

\bibitem{zheng2017dual}
Zheng, Z., Zheng, L., Garrett, M., Yang, Y., Shen, Y.D.: Dual-path
  convolutional image-text embedding.
\newblock arXiv preprint arXiv:1711.05535  (2017)

\end{thebibliography}


\end{document}